%% file: ms.tex
\documentclass[twoside]{article}

\usepackage[accepted]{aistats2021_arxiv}
\usepackage[round]{natbib}

\usepackage{hyperref}       
\usepackage{booktabs}       
\usepackage{algorithm}
\usepackage{algorithmic}
\usepackage{amssymb}
\usepackage{amsmath}
\usepackage{graphicx}
\usepackage{subcaption}
\usepackage{wrapfig}	

\include{defs}

\begin{document}

%

%

\twocolumn[


\aistatstitle{Neural Online Graph Exploration}

\aistatsauthor{ Ioannis Chiotellis \And Daniel Cremers}

\aistatsaddress{ Technical University of Munich \And  Technical University of Munich }
\vspace{-.76cm}
\aistatsaddress{ john.chiotellis@tum.de \And cremers@tum.de}
]

\begin{abstract}
Can we \emph{learn} how to explore unknown spaces efficiently? To answer this question, we study the problem of Online Graph Exploration, the online version of the Traveling Salesperson Problem. We reformulate graph exploration as a reinforcement learning problem and apply Direct Future Prediction~\citep{dosovitskiy2016learning} to solve it. As the graph is discovered online, the corresponding Markov Decision Process entails a dynamic state space, namely the observable graph and a dynamic action space, namely the nodes forming the graph's frontier. To the best of our knowledge, this is the first attempt to solve online graph exploration in a data-driven way. We conduct experiments on six data sets of procedurally generated graphs and three real city road networks. We demonstrate that our agent can learn strategies superior to many well known graph traversal algorithms, confirming that exploration can be learned.
\end{abstract}

\section{INTRODUCTION}

In online graph exploration, an agent is immersed in a completely unknown environment. Located at a node of an unknown graph, they can only see the node's immediate neighbors. The agent moves and discovers the graph as they go. Whenever they visit a new node, all incident edges are revealed, along with their weights and their end nodes. To visit a new node, the agent has to traverse a path of known edges in the discovered graph. For each of these edges, the agent pays their weight as a cost. The goal of the agent is to visit all nodes in the graph, while paying the minimum cost.

By removing any particular geometric constraints, a large number of problems can be reduced to online graph exploration, as it basically is \textit{search} with partial information in a discrete space.
We revisit online graph exploration for undirected unweighted connected graphs. This task can be directly associated with many major problems in robotics such as planning, navigation, tracking and mapping \citep{yamauchi1997frontier}. All these subfields have been thoroughly investigated and a large number of algorithms have been devised, both classical and recently also learning-based. While path planning and navigation algorithms consider the question \textit{``How can I get from A to B the fastest?''}, exploration algorithms consider a more abstract question: \textit{``Where should I go beginning from A in order to discover the world the fastest?''}. In other words, while path planning studies \textit{how} to reach a given destination, exploration is concerned with \textit{which} destination should be reached next, a problem more akin to dynamic planning. We argue that exploration is a fundamental sequential decision making problem and it is therefore worth investigating if algorithms can \textit{learn} which destinations are worth reaching and when.

The best known exploration strategy remains a simple greedy method - the nearest neighbor algorithm (NN). However, as NN selects the nearest (in terms of shortest path distance) unexplored node, its decisions are optimal only when considering a horizon of a single decision step. Therefore, a reasonable question is whether there are algorithms that can consider a longer horizon and thus minimize the \textit{cumulative} path length, which is the true objective of exploration.

We present a learning algorithm that does exactly this. Our contributions can be summarized as follows:
\begin{itemize}
	\item reformulate online graph exploration as a reinforcement learning problem,
	
	\item propose a neural network that can handle the associated dynamic state and action space,
	
	\item show experimentally that the proposed approach solves graph exploration as fast or faster than many classical graph exploration algorithms.
\end{itemize}

\section{RELATED WORK}

The problem of graph exploration has been studied by the graph theory community for decades. A large number of works has been conducted, studying the problem for specific classes of graphs \citep{miyazaki2009online, higashikawa2014online}, with multiple collaborative agents \citep{dereniowski2013fast} or for variations of the problem with  additional information \citep{dobrev2012online} or energy constraints \citep{duncan2006optimal, das2015collaborative}. Successful exploration algorithms are often sophisticated variations of depth first search (DFS), where the algorithm has to decide when to diverge from DFS \citep{kalyanasundaram1994constructing, megow2012online}. A similar problem has been studied by \cite{deng1999exploring} for unweighted \textit{directed} graphs, where the agent has to traverse all edges instead of visiting all nodes. In this setting, the offline equivalent problem is known as the Chinese Postman Problem (CPP) \citep{guan1962graphic} which is solvable in polynomial time. In contrast, the offline equivalent of Online Graph Exploration is the Traveling Salesperson Problem, an NP-hard problem.

Besides the large volume of research, for general graphs, the best known exploration algorithm remains a simple greedy method - the nearest neighbor algorithm (NN). The trajectories followed by NN are provably at most $O(\log n)$\footnote{where $n$ is the number of nodes} longer than the optimal ones \citep{rosenkrantz1977analysis}.

\begin{figure}[h]
	\vspace{.3in}
	\includegraphics[width=\linewidth]{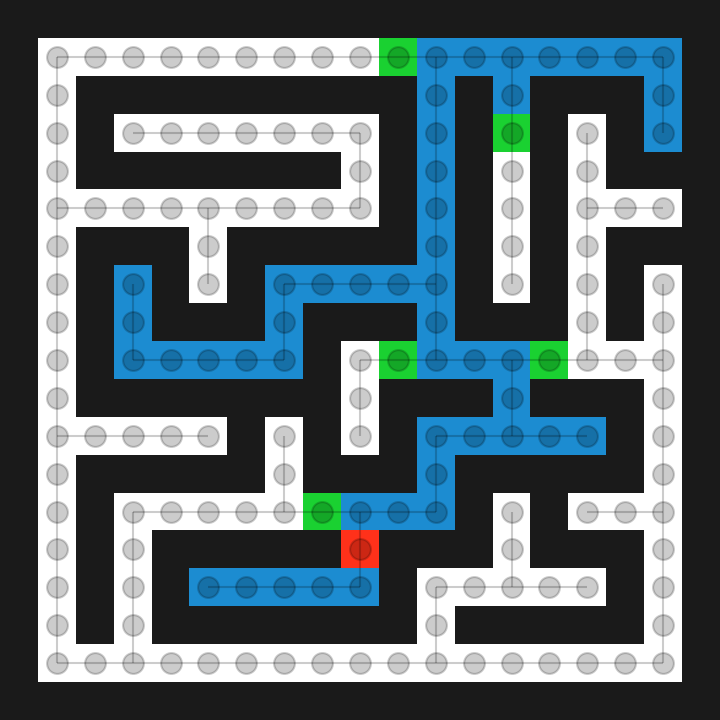}
	\vspace{-.1in}
	\caption{A graph exploration agent (\textbf{red}) keeps track of the nodes already visited (\textbf{blue}) and the \textit{frontier} nodes that could be visited next (\textbf{green}). The frontier nodes form the boundary between known and unknown (\textbf{white}) space. The goal is to discover and visit all nodes as fast as possible.}
	\label{fig:teaser}
\end{figure}

However, to the best of our knowledge, it has not been attempted to solve graph exploration in a data-driven way. In this work, we investigate whether, given a training set of graphs, a learning algorithm is able to find good exploration strategies that are competitive or even superior to traditional methods. Recently, there has been growing interest in applying learning to combinatorial optimization problems. In the work of \cite{vinyals2015pointer}, three problems were studied and solved by learning to ``point'' to elements of a set with a neural network. Among these problems was the Traveling Salesperson Problem, which can be thought of as the offline equivalent to graph exploration. \cite{vinyals2015pointer} proposed a recurrent neural network (RNN) architecture, called Pointer Network, based on the attention mechanism introduced by \cite{bahdanau2015neural}.
However, using an RNN might introduce a bias due to the ordering of the elements in the input sequence. To alleviate this bias, several works have studied neural architectures that can preserve the permutation-invariance property of sets \citep{edwards2016towards,zaheer2017deep, lee2018set}. Moreover, in recent years, Graph Neural Networks (GNNs) \citep{battaglia2018relational} have emerged. These neural networks consider not just sets of nodes but also their pairwise connections or relationships as inputs and can learn how to solve problems such as node classification \citep{kipf2016semi} and link prediction \citep{zhang2018link}. Further, methods such as DeepWalk \citep{perozzi2014deepwalk} and node2vec \citep{grover2016node2vec} have been used to learn node embeddings in an unsupervised way, borrowing ideas from natural language processing \citep{mikolov2013distributed}. These methods make an implicit assumption that nodes that co-appear in a random walk on the graph are more similar than nodes that don't.
Nevertheless, it remains a challenge to learn node embeddings for dynamic graphs that are changing while an algorithm makes decisions about the graph's structure. 

Close to our work is the work of \cite{dai2019learning} which studies exploration on environments with graph-structured state-spaces, such as software testing. In contrast to \citep{dai2019learning}, we study the original problem as defined by the graph theory community which prohibits the revisiting of past nodes. This allows us to study the problem in isolation, without the interference of other tasks such as visual feature learning and the perceptual aliasing problem.


\section{FORMULATION}
\label{sec:formulation}

\begin{figure*}[t!]
	\begin{center}
		\includegraphics[width=\textwidth]{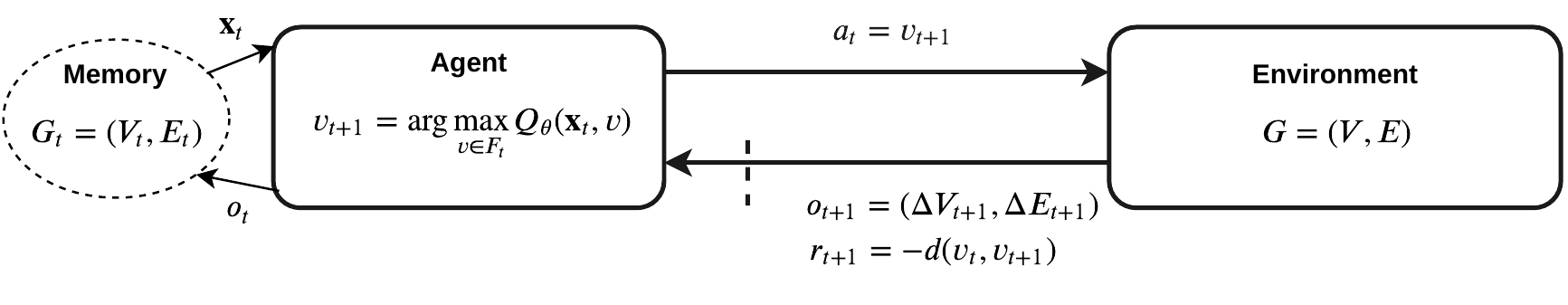}
		\caption{The task of a graph exploring agent is to select the frontier node $v_{t+1} \in F_t$ to visit next. Once the node is visited, the environment reveals a new set of nodes $\Delta V_{t+1}$ and edges $\Delta E_{t+1}$, expanding the agent's knowledge to the graph $\KnownGraph_{t+1} = (V_{t+1}, E_{t+1})$. The distance traveled by the agent from $v_{t}$ to $v_{t+1}$ is paid as a negative reward $r_{t+1}$. The agent acts upon an integrated observation $\intobs_{t+1}$ retrieved from its memory. The subset of nodes that have been discovered but are not yet visited are labeled as the frontier $F_{t+1}$.}
		\label{fig:mdp_reduced}
	\end{center}
	\vskip -0.2in
\end{figure*}

\subsection{Graph Exploration Overview}
\label{sec:oge-as-rl}
An agent explores a connected unweighted graph $G=(V, E)$. At time step $t=0$, they start at an arbitrary node $v_0 \in V$ and they can only observe an initial \textit{map} $\KnownGraph_0 = (V_0, E_0)$ comprised of the neighbors and incident edges of $v_0$. We assume the agent has a memory where it can store and integrate observations. Therefore, at any time step $t$, the agent observes a subgraph $\KnownGraph_t = (V_t, E_t)$ with a subset of visited nodes $C_t$ and a subset of frontier nodes $F_t$ that are to be explored.
Being at node $v_t \in C_t$, the agent has to choose a node $v_{t+1} \in F_t$ to visit next. Once a decision is made, the agent follows a path from $v_t$ to $v_{t+1}$ of length $l_{t+1} = d_{\KnownGraph_t}(v_t, v_{t+1})$. Note that this is a shortest path in $\KnownGraph_t$ but not necessarily in $G$\footnote{In the rest of the paper, we omit $\KnownGraph_t$ and use the shorter notation $d(v_t, v_{t+1})$ wherever possible.}. The new node $v_{t+1}$ gets removed from the frontier and becomes part of the set of visited nodes: $C_{t+1} \leftarrow C_t \cup \{v_{t+1}\}$. Finally, the agent observes the neighbors $N(v_{t+1})$ of $v_{t+1}$ and the frontier gets expanded by the subset of neighbors that have not been observed in the past:
\begin{equation}
F_{t+1} \leftarrow F_t \cup N(v_{t+1}) \setminus C_{t+1}.
\end{equation} 

The goal is to visit the nodes in such an \textit{order} that the total path length is minimized. Notice that we use a different timescale than commonly used \eg in navigation problems. In a single time step, the exploration agent can traverse a path of arbitrary length in the known graph $\KnownGraph_t$. 
The differences between exploration algorithms lie in the way they choose the node $v \in F_t$ to visit next. For instance, DFS considers the order of entry in the frontier and chooses the most recently entered node. The nearest neighbor algorithm (NN) chooses the node closest to the current node $v_t \in C_t$:
\begin{equation}
v_{t+1}^{NN} = \argmin_{v \in F_t} d(v_t, v)
\end{equation}

However, since the NN selection rule is greedy, it might be suboptimal. Namely, an algorithm $A$ could exist that takes into account the \textit{expected} future path lengths and could therefore make better decisions:
\begin{equation}
v_{t+1}^A = \argmin_{v \in F_t} \mathbb{E}[d(v_t, v) + \sum_{i=t+1}^{\infty} d(v_i, v_{i+1})].
\end{equation}

This formulation is reminiscent of reinforcement learning (RL). In RL an agent in a state $s_t$ and following a policy $\pi$, chooses action $a_t = \pi(s_t)$ and receives an immediate reward $r_{t+1}$. The true objective of the agent is to maximize the cumulative reward:
\begin{equation}
\label{eq:fixed_action_space}
a_t = \argmax_{a \in \mathcal{A}} \mathbb{E}[r_{t+1} + \sum_{i=t+1}^{\infty} \gamma^{i-t} r_{i+1} | \pi, a],
\end{equation} 

where $\gamma \in [0, 1]$ is a discount factor that weighs distant future rewards less than imminent rewards. The expectation of cumulative rewards is also known as the action value $Q^\pi(s_t, a_t)$. Notice that the NN algorithm can be exactly recovered for $\gamma = 0$.

\subsection{Markov Decision Process}
\label{sec:mdp}
RL problems are formally described as Markov Decision Processes (MDPs). An MDP is defined as a 5-tuple $(\mathcal{S}, \mathcal{A}, p, r, \gamma)$, namely a state space $\mathcal{S}$, an action space $\mathcal{A}$, a state transition probability function $p: \mathcal{S} \times \mathcal{A} \times \mathcal{S} \mapsto [0, 1]$, a reward function $r: \mathcal{S} \times \mathcal{A}\mapsto \mathbb{R}$ and a discount factor $\gamma  \in [0, 1]$. A partially observable Markov decision process (POMDP) is a generalization of a MDP, where the agent cannot directly observe the state $s_t \in \mathcal{S}$ but has partial information through observations $o_t \in \mathcal{O}$. An agent with a memory component can integrate partial observations to cumulative observations $\intobs_t \in \mathcal{X}$. Notice that the observation space $\mathcal{X}$ is a subset of the state space $\mathcal{S}$. Thus, we refer to the setting of Online Graph Exploration as a memory-augmented POMDP. In Figure~\ref{fig:mdp_reduced}, we illustrate this setting. In the following, we describe the components of this MDP.

\paragraph{State Space}
Let $\mathcal{G}$ be the set of all conceivable graphs, and let $\mathcal{P}_G$ denote the set of all conceivable visit orderings $P_G$ for a graph $G\in\mathcal{G}$.  Then the state space $\mathcal{S}$ is defined as the set of all pairs $(G, P_G)$ of graphs $G\in\mathcal{G}$ and associated visit orderings $P_G\in\mathcal{P}_G$:

\paragraph{Observation Space}
At each time step, the environment reveals the neighborhood of the visited node. Therefore, the observation space is exactly the subset of graphs that are star graphs.

\paragraph{Action Space}
It is common in RL problems with discrete action spaces, for the agent to have access to a fixed set of actions $\mathcal{A}$ as in Eq.~\eqref{eq:fixed_action_space}. Instead, in graph exploration, a new unique action set $\mathcal{A}_t$ is induced from the state at each time step. This action set corresponds to the nodes that have been observed but not visited yet, namely the nodes in the frontier:
\begin{equation}
\mathcal{A}_t = F_t.
\end{equation}
The general action space can be described by the power set of all nodes: $\mathcal{A} = 2^\mathcal{V}$.
Note that the frontier can be derived from the known graph $\KnownGraph_t=(V_t, E_t)$ 
and the path $\Path_t$ as $F_t = V_t \setminus C_t$, where $C_t$ denotes the \textit{set} of visited nodes found in the sequence $P_t$.

\paragraph{Reward Function}
As defined in section \ref{sec:oge-as-rl}, the rewards correspond to negative geodesic distances. Therefore, assuming unweighted graphs, all rewards are strictly negative:
\begin{equation}
r(s_t, a_t) = -d(v_t, v_{t+1}) < 0.
\end{equation}

\paragraph{State Transition Function}
Let $v_{t+1}$ be the node to visit next. Then, if $s_t = (G, P_t)$ is the state described by the graph $G$ and the path $P_t$, the new state is described by the same graph $G$ and the extended path $\Path_{t+1} \leftarrow \Path_{t} \; || \; v_{t+1}$\footnote{By $||$ we denote concatenation of a sequence with a new element.}.

\paragraph{Memory Update}
Upon observing $\Delta V_{t+1} = N(v_{t+1})$, namely the neighbors of $v_{t+1}$, and $\Delta E_{t+1} = E(v_{t+1})$, the edges from $v_{t+1}$ to $N(v_{t+1})$, the agent's memory is updated as:
\begin{align}
V_{t+1} &\leftarrow V_t \cup \Delta V_{t+1} \\
E_{t+1} &\leftarrow E_t \cup \Delta E_{t+1} \\
\Path_{t+1} &\leftarrow \Path_{t} \; || \; v_{t+1} \\
C_{t+1} &\leftarrow C_t \cup \{v_{t+1}\}\\
F_{t+1} &\leftarrow F_t \cup \Delta V_{t+1} \setminus C_{t+1}.
\end{align}

\section{METHODOLOGY}
\label{sec:methodology}
\subsection{Predicting the Future Path Lengths}
Our premise is that a learning agent can perform better than traditional exploration algorithms, as long as they can predict the future distances to be traveled. Inspired by the framework introduced by \cite{dosovitskiy2016learning}, we use Direct Future Prediction (DFP) to learn a predictor of future path lengths. Confirming the authors' observations, we found that reducing policy learning to a supervised regression problem makes training faster and more stable. In particular, at time $t$, we aim to predict the vector
\begin{equation}
\targetvec_t = (\measvec_{t+\tau_1} - \measvec_t, \measvec_{t+\tau_2} - \measvec_t, \dots, \measvec_{t+\tau_M} -\measvec_t),
\end{equation}

where $\measvec_t$ is a low-dimensional measurement vector augmenting the agent's high-dimensional observation $\intobs_t$, and $\{\tau_j\}_{j=1}^M$ are temporal offsets. Following \cite{dosovitskiy2016learning}, we choose exponential offsets $\tau_j = 2^{j-1}$. We could directly use a scalar measurement $L_t = \sum_{i=0}^{t} l_i$, namely the cumulative path length up to time $t$. However, there are several disadvantages with this choice. For unweighted graphs, we know that any \textit{one-step} path length $l_t$ lies in the range $[1, N_{max}-1]$, where $N_{max}$ is the maximum number of nodes we are considering. Thus, directly predicting path lengths would limit our ability to generalize to graphs larger than our training graphs. 
Second, the distribution of path lengths naturally grows over time together with the observable graph's diameter. To avoid these problems, instead of minimizing path lengths, we maximize the agent's exploration rate $u_t = \frac{|C_t|}{L_t} = \frac{t}{L_t}$ which always lies in the $[0, 1]$ interval and thus the entries of $\targetvec$ always lie in $[-1, 1]$. At test time, we choose the node to visit next by simply taking the $\argmax$:
\begin{equation}
v_{t+1} = \argmax_{v \in F_t} \mathbf{g}^\top \prednet (\intobs_t, \measvec_t, v),
\end{equation}

where $\prednet$ is our parameterized predictor network, $\intobs_t=(V_t, E_t, X_t)$ is the observable graph with node features $X_t$ and $\mathbf{g}$ is a goal vector expressing how much we care about different future horizons. 
Another advantage of using DFP instead of RL is that, given information about the remaining time available, we can directly incorporate it in the goal vector, both at training and at test time without the need of retraining the network. In contrast to \cite{dosovitskiy2016learning}, we don't use $\mathbf{g}$ as an input to the network~\footnote{We found that adding a goal module does not improve and some times even hurts performance.}, but only as a weighting of the predictions to obtain a policy.

\begin{figure*}[t!]
	\begin{center}
		\includegraphics[width=\linewidth, height=.33\textheight]{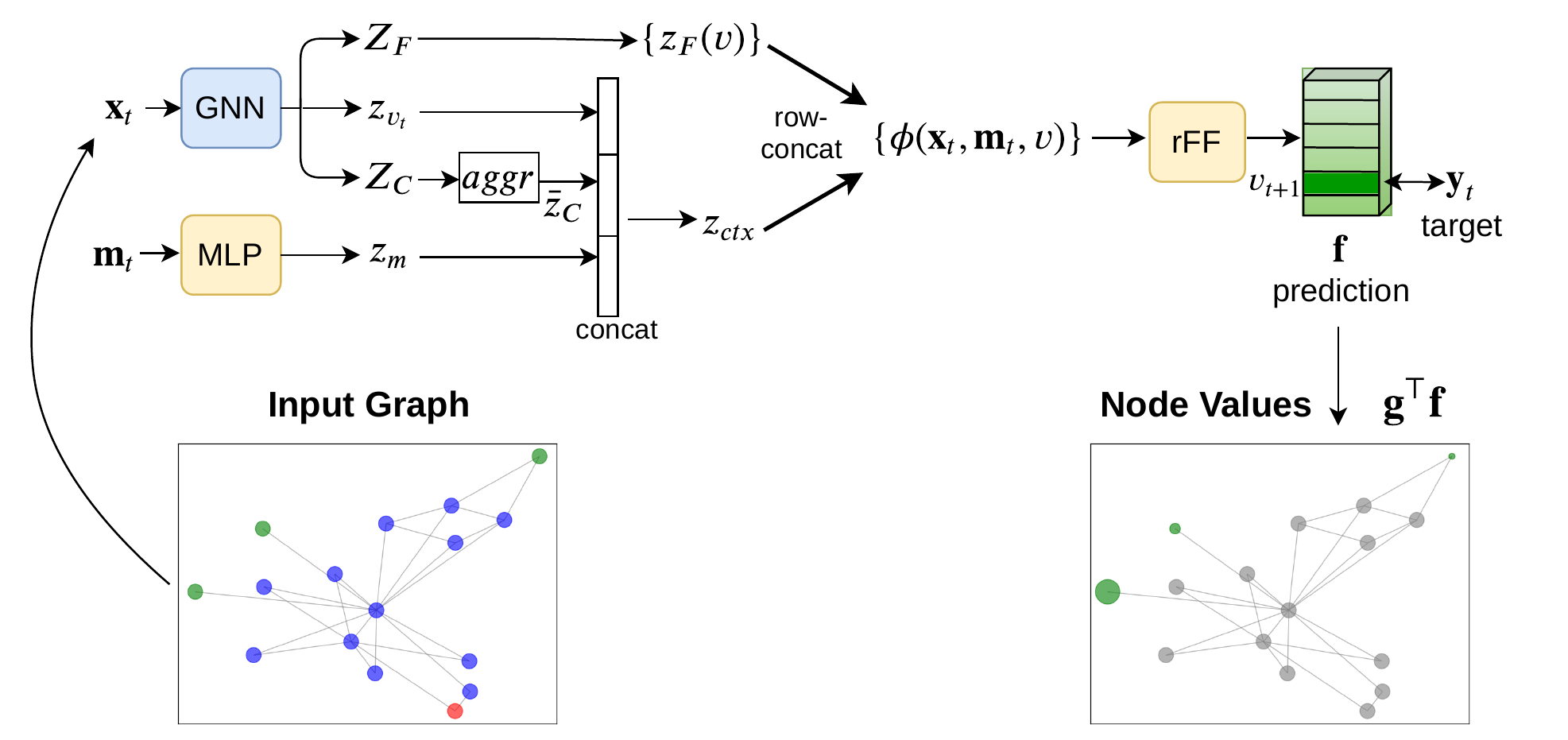}
		\caption{The proposed neural network architecture for online graph exploration.}
		\label{fig:dots_net}
	\end{center}
	\vskip -0.2in
\end{figure*}

\subsection{Network Architecture}
\label{sec:implementation}

In Figure~\ref{fig:dots_net} we show our network architecture. We first obtain node embeddings $Z_F$, $Z_C$ and $z_{v_t}$ by passing the observable graph $\intobs_t=(V_t, E_t, X_t)$ through a graph neural network (GNN). The embeddings correspond to the node subsets $F_t$, $C_t$ and the current node $v_t$. We use a standard graph convolutional network (GCN) \citep{kipf2016semi}, but any graph neural network that produces node embeddings can be used. We then aggregate the node embeddings $Z_C$ of the visited set to obtain a subgraph embedding $\bar{z}_{C}$. Even though more sophisticated pooling methods (\eg attention~\citep{velivckovic2017graph}) may be used, we use simple mean pooling. 

We pass $\measvec_t$ through a multi-layer Perceptron (MLP) to obtain a measurement embedding $z_m$. The vectors $\bar{z}_C$, $z_{v_t}$ and $z_m$ are concatenated to form a \emph{context} vector $z_{ctx}$. Each frontier node embedding $z_F(v) \in Z_F$ is concatenated with $z_{ctx}$, resulting in a set of state-action encodings $\phi(\intobs_t, \measvec_t, v)$, one for each node $v \in F_t$. Finally we pass these encodings through a row-wise feed-forward network (another MLP) to obtain a prediction vector $\predvec(\intobs_t, \measvec_t, v)$ for each node $v \in F_t$. The state-action value $Q(\intobs_t, \measvec_t, v, \mathbf{g})$ of each node in the frontier can be obtained by multiplying the associated prediction vector $\predvec(\intobs_t, \measvec_t, v)$ with the goal vector $\mathbf{g}$.

\subsection{Input Features}
\label{sec:input-features}

Solving online graph exploration as a learning problem depends critically on what the learning algorithm ``sees'' as input. The state $s_t$ consists of the known graph $\KnownGraph_t$ and the path $\Path_t$. Therefore, we have to decide on the nodes and perhaps also edge input features of the graph $\KnownGraph_t$. Second, we have to consider a representation of the path $\Path_t$. 

We used categorical node features indicating if a node belongs to the visited set $C_t$, the frontier set $F_t$ or if it is the current node $v_t$:
\begin{equation}
x(v_i) = [\mathbf{1}(v_i \in C_t), \mathbf{1}(v_i \in F_t), \mathbf{1}(v_i = v_t)].
\end{equation}

Note that a categorical node feature space can incorporate many classical graph traversal algorithms (\eg BFS, DFS and NN) as special cases by simply adding a binary channel that indicates the node that would be selected by the respective algorithm. Furthermore, this representation allows the learning algorithm to potentially learn a \textit{hyper-policy} \citep{precup1998theoretical} by combining greedy algorithms in novel ways. 

In preliminary experiments, we investigated ways to utilize the order of visit of the nodes, $P_t$, by using positional encodings \citep{vaswani2017attention} as continuous node features. We found that these features degraded the agent's performance both when used on their own and when combined with the categorical features.

\subsection{Training}
In Algorithm~\ref{alg:noge-training} we describe our training procedure. In each episode, we randomly sample a graph from the training set and then randomly set one of its nodes as source. This virtually increases the training set size from the number of training graphs $|\mathcal{G}_{train}|$ to the total number of nodes in all training graphs $\sum_{G \in \mathcal{G}_{train}} |V_G|$, where by $V_G$ we denote the vertices of graph $G$.

\begin{algorithm}[t!]
	\caption{Training \text{\Method}}
	\label{alg:noge-training}
	\begin{algorithmic}[1]
		\STATE {\bfseries Input:} network $\prednet$, training set of graphs $\mathcal{G}_{train}$, time limit $T_{max}$, goal vector $\goalvec$, minibatch size B.
		\STATE {\bfseries Output:} trained network $\prednet$.
		\STATE Initialize $\theta$ randomly.
		\STATE Initialize an experience replay buffer $\mathcal{R}$.
		\WHILE{\emph{training}}
		\STATE Sample a graph $G=(V, E) \sim \mathcal{G}_{train}$.
		\STATE Sample a source node $\SourceNode_0 \sim V$.
		\STATE Explore $G$
		using $\prednet$,  $\goalvec$ and $\epsilon$-greedy policy for up to $T_{max}$ episode steps.  \\
		\STATE Store $G_T$ and the followed path $P_T$ in $\mathcal{R}$.
		\STATE Sample a minibatch $\{G_t^{(i)}, P_t^{(i)}\}_{i=1}^B$ from $\mathcal{R}$.
		\STATE Reconstruct tuples $\{\intobs_t^{(i)}, \measvec_t^{(i)}, v_{t+1}^{(i)}, \targetvec_t^{(i)}\}_{i=1}^B$.
		\STATE Train $\prednet$ with the minibatch, using Adam \citep{kingma2014adam} to minimize the mean squared error:\\
		$L(\theta) = \frac{1}{B} \sum_{i=1}^{B} || \targetvec_t^{(i)} - \prednet(\intobs_t^{(i)}, \measvec_t^{(i)}, v_{t+1}^{(i)})||^2$.
		\ENDWHILE
	\end{algorithmic}
\end{algorithm}

\section{EXPERIMENTS}
\label{sec:results}

The complete set of hyperparameters used is reported in Appendix~\ref{appendix:impl} and our full source code is available at \url{https://github.com/johny-c/noge}.

\subsection{Evaluation Protocol}
We evaluate our algorithm - \text{\Method} (Neural Online Graph Exploration) on data sets of generated and real networks. In addition to the basic version of our algorithm, we  evaluate NOGE with an extra node feature, indicating the nearest neighbor, as described in section~\ref{sec:input-features}. We call this variant NOGE-NN. We use three well known graph exploration algorithms as baselines: Breadth First Search (BFS), Depth First Search (DFS) and Nearest Neighbor (NN). We note that these heuristics do not need any training. For completeness, we also report a random exploration baseline (RANDOM). We compare the algorithms in terms of the \textit{exploration rate} $u_T$, namely the number of visited nodes over the total path length at the end of episodes:

\begin{equation}
u_T = \frac{|C_T|}{|P_T|} = \frac{T}{\sum_{i=0}^{T} l_i}.
\end{equation}

For the test sets we fix a set of source nodes per graph, to compare all methods given the exact same initial conditions. The metrics reported are computed on the test sets after either all nodes have been explored or when a fixed number of $T_{max}=500$ exploration steps has been reached. For all experiments we report mean and standard deviation over 5 random seeds.

\subsection{Procedurally Generated Graphs}
\label{experiment:synth}

\begin{figure*}[h!]
		\vskip -0.1in
	\begin{center}
		\includegraphics[height=0.31\textheight]{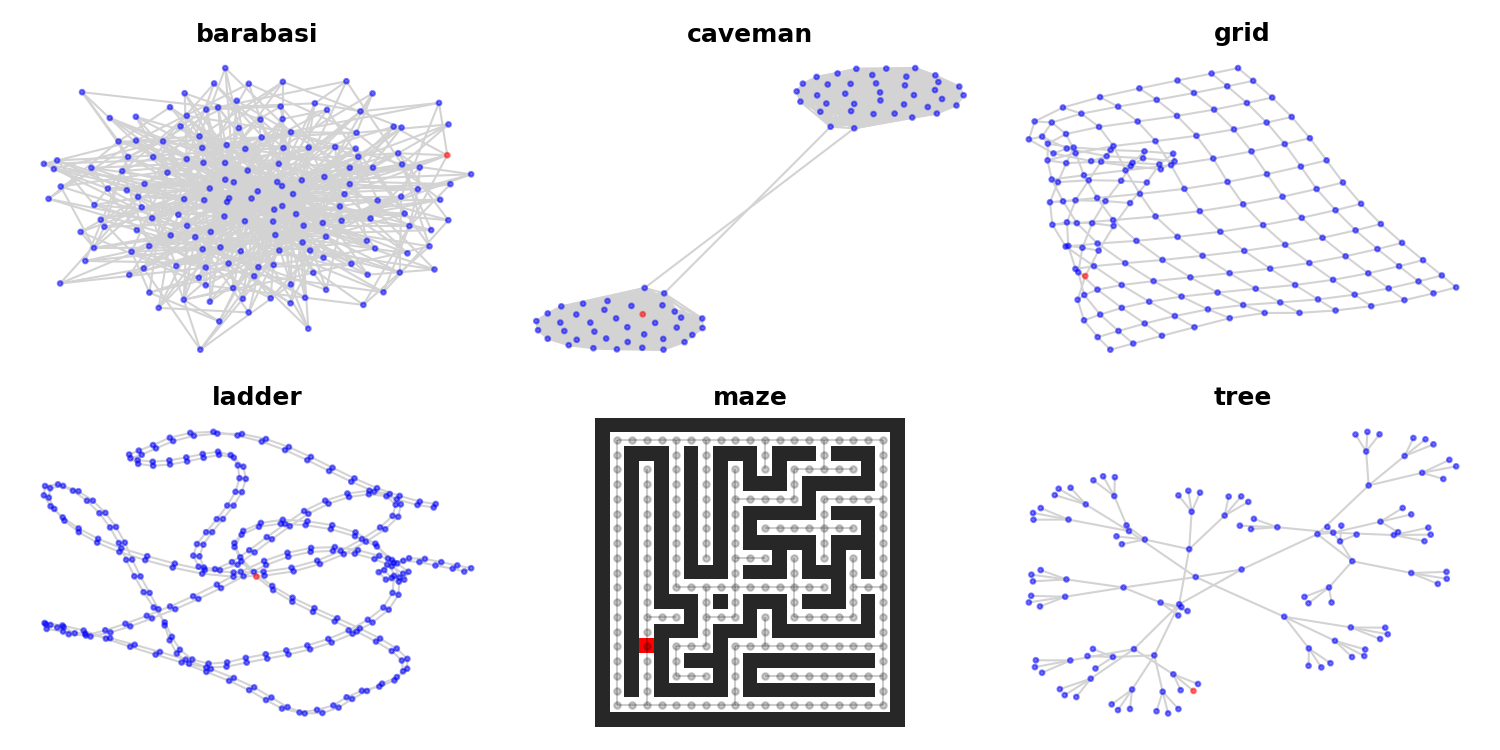}
		\caption{Samples from the procedurally generated data sets used in the experiments of Section~\ref{experiment:synth}.}
		\label{fig:data_synth}
	\end{center}
	\vskip -0.2in
\end{figure*}

We first examine six classes of procedurally generated graphs (Figure~\ref{fig:data_synth}). We used the \textit{networkx} library (\url{https://github.com/networkx/networkx}) to generate a diverse (in terms of size and connectivity) set of graphs for each class. In Table~\ref{tbl:data-synth-stats} we report basic statistics describing the size and connectivity of the graphs. We split each data set in a training (80\%) and test set (20\%) of graphs. For some of these data sets, an optimal strategy is known. For instance, DFS explores trees optimally by traversing each edge two times - once to explore and once to backtrack. Thus its exploration rate is approximately $0.5$. It is worthwhile examining if \text{\Method} can find this optimal strategy.

\begin{figure*}[h!]
	\vspace{.1in}
	\begin{center}
		\includegraphics[width=\linewidth, height=0.4\textheight]{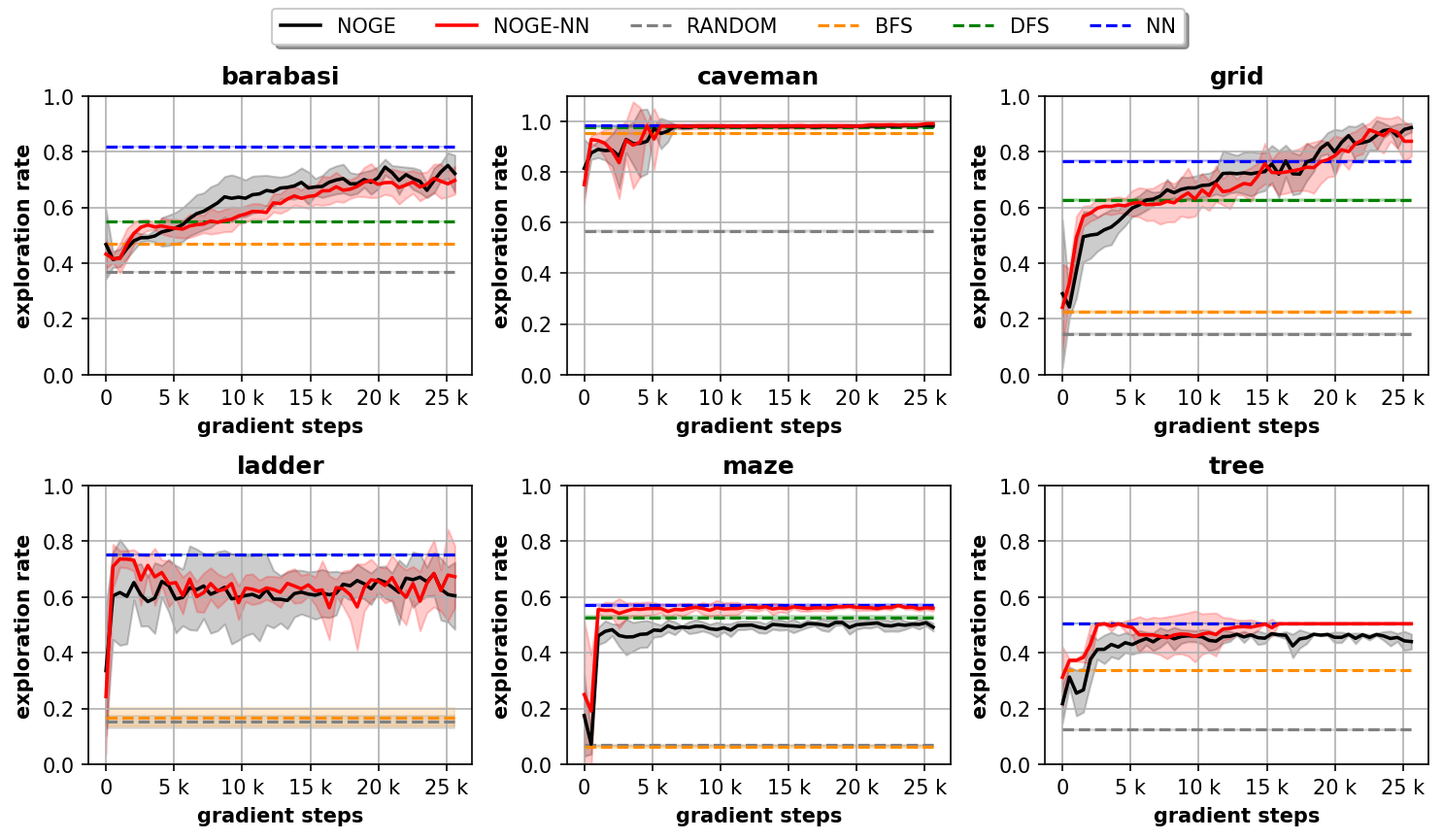}
			\caption{Exploration rate over gradient steps for the six procedurally generated data sets.}
			\label{fig:curves-synth}
	\end{center}
\end{figure*}

In Figure~\ref{fig:curves-synth} we show the test performance of NOGE over 25600 training steps.  In Table~\ref{tbl:synth-final-results} we report the final performance of the algorithms compared to the baselines. NOGE is able to outperform other methods on grids and the caveman data set and find the optimal strategy on trees. Somewhat surprisingly the NN feature seems to only help on the maze and tree data sets. Note that in ladder and tree, DFS's line is hidden as its performance matches that of NN.

\begin{table}[h!]
	\caption{Basic statistics of the procedurally generated data sets (tr: training set, te: test set).}
			\vspace{-.1in}
	\begin{center}
		\input{tables/synth_data_stats.tex}
	\end{center}
	\label{tbl:data-synth-stats}
\end{table}

\begin{table}[b!]
	\caption{Basic statistics of the city road network data sets (tr: training set, te: test set).}
	\vspace{-.1in}
	\begin{center}
		\input{tables/real_data_stats.tex}
	\end{center}
	\label{tbl:data-real-stats}
\end{table}

\begin{table*}[t!]
	\caption{Final exploration rate: Mean and standard deviation on the generated data sets.}
	\vspace{-.15in}
	\begin{center}
		\small
		\input{tables/synth_results.tex}
	\end{center}
	\label{tbl:synth-final-results}
\end{table*}

\subsection{City Road Networks}
\label{experiment:real}
\begin{figure*}[h!]
	
	\begin{subfigure}{0.32\linewidth}
		\includegraphics[height=.16\textheight]{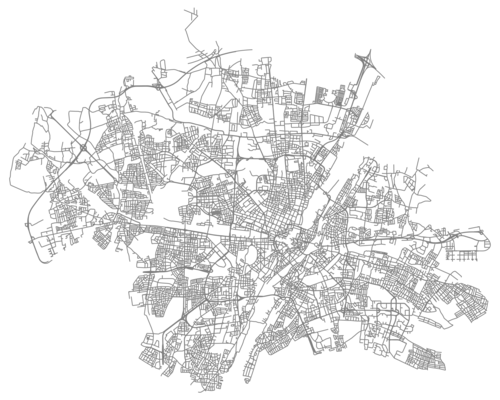} 
		\caption{Munich}
		\label{fig:muc}
	\end{subfigure}
	\begin{subfigure}{0.32\linewidth}
		\includegraphics[height=.16\textheight]{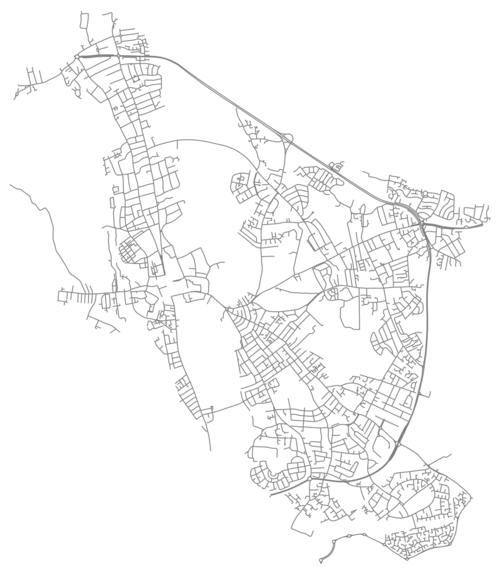}
		\caption{Oxford}
		\label{fig:oxf}
	\end{subfigure}
	\begin{subfigure}{0.32\linewidth}
		\includegraphics[height=.16\textheight]{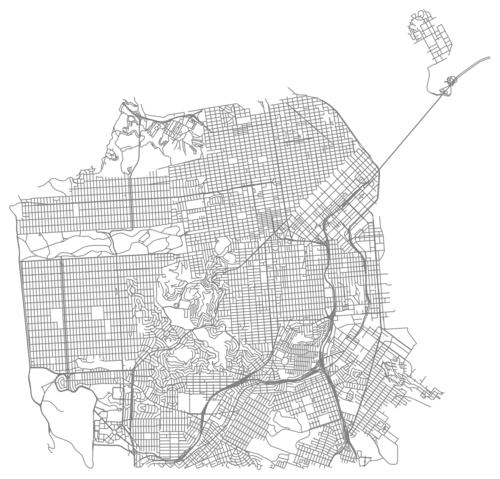}
		\caption{San Francisco}
		\label{fig:sfo}
	\end{subfigure}
	
	\caption{The road city networks used for the experiments in Section~\ref{experiment:real}.}
	\label{fig:data-real}
\end{figure*}

\vspace{-.15cm}
To examine its capabilities, we also evaluate \text{\Method},  on three real road networks. We use openly available drivable road networks from OpenStreetMap \citep{haklay2008openstreetmap}. We explore three cities with diverse networks: Munich (MUC), Oxford (OXF) and San Francisco (SFO), shown in Figure \ref{fig:data-real}.

In Table~\ref{tbl:data-real-stats}, we show basic statistics for these data sets. We constructed a training set and test set for each city by cutting each graph in two components and removing any edges connecting them. The cut was defined by the diagonal of each city's 2D bounding box. The larger component (approx. 60\% of the nodes) was used for training and the smaller one for testing.

\begin{figure*}[h!]
	\begin{center}
		\includegraphics[width=\linewidth, height=0.25\textheight]{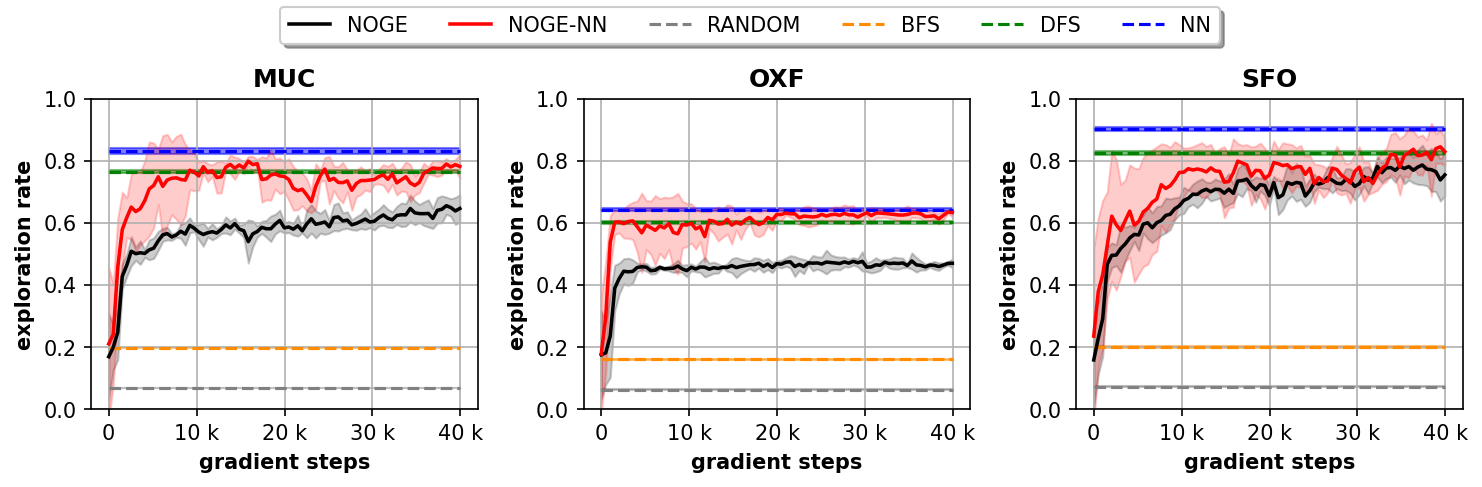}
			\caption{Exploration rate over gradient steps for the city road networks data sets.}
		\label{fig:curves-real}
	\end{center}
	\vspace{-.1in}

\end{figure*}

In Figure~\ref{fig:curves-real} we show the test performance of NOGE over 40000 training steps and in Table~\ref{tbl:real-final-results} we report the final performance. Nearest Neighbor performs clearly better in San Francisco and Munich. NOGE-NN is able to match and surpass DFS, as the second best method. In Oxford, NOGE-NN is within a standard deviation from the best performance by NN. In these graphs the NN feature clearly improves performance.

\begin{table*}[t!]
	\caption{Final exploration rate: Mean and standard deviation on the city road networks data sets.}
		\vspace{-.1in}
	\begin{center}
		\small
		\input{tables/real_results.tex}
	\end{center}
	\label{tbl:real-final-results}
\end{table*}

\section{CONCLUSION}
\label{sec:conclusion}

In this work, we presented \text{\Method}, a learning-based algorithm for exploring graphs online. First, we formulated an appropriate memory-augmented Markov Decision Process. Second, we proposed a neural architecture that can handle the growing graph as input and the dynamic frontier as output. Third, we devised a node feature space that can represent greedy methods as \emph{options}~\citep{precup1998theoretical}. Finally, we showed experimentally that \text{\Method} is competitive to well known classical graph exploration algorithms in terms of the exploration rate of unseen graphs.

\nocite{mnih2015human}
\newpage
\bibliographystyle{apalike}
\bibliography{references2}

\newpage

\appendix

\include{appendix}

\end{document}

%% file: defs.tex
\newcommand{\SourceNode}{v}

\newcommand{\KnownGraph}{G}
\newcommand{\Path}{P}
\newcommand{\Method}{\text{NOGE}} 

\newcommand{\eg}{\textit{e.g. }}

\newcommand{\prednet}{\mathbf{f}_{\theta}}
\newcommand{\targetvec}{\mathbf{y}}
\newcommand{\predvec}{\mathbf{f}}
\newcommand{\measvec}{\mathbf{m}}
\newcommand{\intobs}{\mathbf{x}}
\newcommand{\goalvec}{\mathbf{g}}

\DeclareMathOperator*{\argmax}{arg\,max}
\DeclareMathOperator*{\argmin}{arg\,min}

%% file: tables/synth_data_stats.tex
\begin{tabular}{lccc}
\toprule
      Dataset &  Size & $|V|_{min,max}$ & $|E|_{min,max}$ \\
\midrule
 barabasi(tr) &   400 &      (100, 199) &      (384, 780) \\
 barabasi(te) &   100 &      (100, 199) &      (384, 780) \\
   ladder(tr) &    80 &      (200, 398) &      (298, 595) \\
   ladder(te) &    20 &      (220, 386) &      (328, 577) \\
     tree(tr) &     4 &     (121, 1365) &     (120, 1364) \\
     tree(te) &     2 &     (364, 1093) &     (363, 1092) \\
     grid(tr) &    80 &       (64, 289) &      (112, 544) \\
     grid(te) &    20 &       (72, 240) &      (127, 449) \\
  caveman(tr) &   120 &       (60, 316) &    (870, 12324) \\
  caveman(te) &    30 &       (70, 304) &   (1190, 11400) \\
     maze(tr) &   400 &       (97, 251) &       (96, 262) \\
     maze(te) &   100 &       (97, 255) &       (96, 276) \\
\bottomrule
\end{tabular}

%% file: tables/real_data_stats.tex
\begin{tabular}{lccc}
\toprule
 Dataset &  Size & $|V|_{min,max}$ & $|E|_{min,max}$ \\
\midrule
 MUC(tr) &     1 &    (8559, 8559) &  (12821, 12821) \\
 MUC(te) &     1 &    (5441, 5441) &    (7772, 7772) \\
 OXF(tr) &     1 &    (2197, 2197) &    (2561, 2561) \\
 OXF(te) &     1 &    (1185, 1185) &    (1430, 1430) \\
 SFO(tr) &     1 &    (5691, 5691) &    (9002, 9002) \\
 SFO(te) &     1 &    (3885, 3885) &    (6579, 6579) \\
\bottomrule
\end{tabular}

%% file: tables/synth_results.tex
\begin{tabular}{lrrrrrr}
\toprule
  Dataset &           RANDOM &              BFS &              DFS &               NN &             NOGE &          NOGE-NN \\
\midrule
 barabasi &  0.3695 (0.0006) &  0.4695 (0.0013) &  0.5494 (0.0009) &  $\mathbf{0.8179 (0.0014)}$ &  0.7214 (0.0663) &  0.6970 (0.0497) \\
  caveman &  0.5664 (0.0050) &  0.9526 (0.0025) &  0.9778 (0.0006) &  0.9827 (0.0015) &  0.9817 (0.0029) &  $\mathbf{0.9907 (0.0031)}$ \\
     grid &  0.1461 (0.0037) &  0.2264 (0.0039) &  0.6272 (0.0041) &  0.7670 (0.0028) &  $\mathbf{0.8861 (0.0162)}$ &   $\mathbf{0.8373 (0.0536)}$ \\
   ladder &  0.1531 (0.0226) &  0.1691 (0.0341) &  $\mathbf{0.7519 (0.0010)}$ &  $\mathbf{0.7530 (0.0009)}$ &  0.6046 (0.1208) &  $\mathbf{0.6729 (0.1114)}$ \\
     maze &  0.0688 (0.0027) &  0.0626 (0.0025) &  0.5266 (0.0050) &  $\mathbf{0.5723 (0.0033)}$ &  0.4921 (0.0140) &  0.5601 (0.0106) \\
     tree &  0.1242 (0.0011) &  0.3397 (0.0002) &  $\mathbf{0.5044 (0.0002)}$ &  $\mathbf{0.5044 (0.0003)}$ &  0.4403 (0.0272) &  $\mathbf{0.5043 (0.0004)}$ \\
\bottomrule
\end{tabular}

%% file: tables/real_results.tex
\begin{tabular}{lcccccc}
\toprule
Dataset &           RANDOM &              BFS &              DFS &               NN &             NOGE &          NOGE-NN \\
\midrule
    MUC &  0.0674 (0.0003) &  0.1961 (0.0007) &  0.7644 (0.0053) &  $\mathbf{0.8314 (0.0091)}$ &  0.6458 (0.0441) &  0.7814 (0.0386) \\
    OXF &  0.0624 (0.0009) &  0.1608 (0.0019) &  0.6012 (0.0048) &  $\mathbf{0.6422 (0.0037)}$ &  0.4695 (0.0136) &  $\mathbf{0.6328 (0.0141)}$ \\
    SFO &  0.0726 (0.0015) &  0.2007 (0.0033) &  0.8252 (0.0073) &  $\mathbf{0.9017 (0.0064)}$ &  0.7541 (0.0679) &  0.8289 (0.0456) \\
\bottomrule
\end{tabular}

%% file: appendix.tex
\section{IMPLEMENTATION DETAILS}
\label{appendix:impl}

\subsection{Hyperparameters}
In Table~\ref{tbl:synth-cfg} we show the hyperparameters used for the experiments on the procedurally generated data sets. The only differences in the experiments for the city road networks are the number of training steps which was set to 40000 and the hidden layer width of the neural network (see next subsection). We elaborate on hyperparameters, the usage of which may not be immediately clear:

\paragraph{Node History} 
It is common in deep reinforcement learning to replace the input observation $\intobs_t$ with a stack of the last $k$ observations $[\intobs_{t-k+1}, \intobs_{t-k+2}, \ldots, \intobs_t]$, particularly when the observations are images. This gives the agent a sense of the environment dynamics. We found that using a stack of the last 2 feature vectors for each node also improves performance in graph exploration, as it gives a sense of direction. 

\paragraph{Feature Range} 
As a preprocessing step, shifting input features to the $[-0.5, 0.5]$ range speeds up learning.

\paragraph{Target Normalization}  
As a postprocessing step, target normalization also aids the learning process. We scaled targets $\mathbf{y}$ by the standard deviation of measurements collected during random exploration, as described by \cite{dosovitskiy2016learning}.

\paragraph{Evaluation Episodes}
For evaluation, we sampled 50 graphs from the test set and fixed one source node per graph. If the test set contained less than 50 graphs, we sampled 50 source nodes uniformly from all test graphs.

\paragraph{$\varepsilon$-greedy Policy}
As described in our training algorithm, we used an $\varepsilon$-greedy policy to collect experiences, namely a random frontier node was selected to be visited with probability $\varepsilon$ and a node was selected by the network's policy with probability 1 - $\varepsilon$. We linearly interpolated $\varepsilon$ from 1 to 0.15 over the course of training. During testing, the greedy policy ($\varepsilon=0$) was used.

\begin{table}[h!]
	\caption{Hyperparameters used in experiments on procedurally generated graphs.}
	\vspace{-.1in}
	\begin{center}
		\input{tables/cfg_synth.tex}
	\end{center}
	\label{tbl:synth-cfg}
\end{table}

\subsection{Network Architecture}
The architecture of our network, used for the procedurally generated graphs, is shown in Table~\ref{tbl:net-synth-cfg}. The same architecture was used for the city road networks except that all layers - apart for input and output - are wider by a factor of two. The input dimension for the graph neural network (GNN) was 3 for NOGE and 4 for NOGE-NN. In all networks we use the ReLU nonlinearity after all layers except for the output layer of the row-wise feed-forward (rFF) network.

\subsection{Replay Buffer for Graphs}
To use the replay buffer for training, we need to be able to sample graph observations $G_t$ from any time step in an episode. To do that, for each episode we store the discovered graph $G_T=(V_T, E_T)$ at the end of the episode and two arrays: an array of node counts and an array of edge counts, indicating the size of the graph at each time step. To be able to reconstruct the frontier at an arbitrary time step $t$, we need to store two integers per node $v$: the time of discovery $t_{dis}(v)$ and the time of visit $t_{dis}(v)$. Then the frontier $F_t$ at any time step $t$ is:
\begin{equation}
	F_t = \{v \in V_t : t \geq  t_{dis}(v) \land t < t_{vis}(v) \}
\end{equation}

\begin{table}[h!]
	\caption{The network architecture.}
	\vspace{-.1in}
	\begin{center}
		\begin{tabular}{lcc}
			\toprule
			module & input dimension & output dimension \\
			\midrule
			GNN    &			   3 or 4 & 32 \\
				   &			  32 & 64 \\
			\midrule
			MLP	   &			   1 & 64 \\
				   &			  64 & 64 \\
			\midrule
			rFF    &			 256 & 128 \\
				   &			 128 & 8 \\
			\bottomrule
		\end{tabular}
	\end{center}
	\label{tbl:net-synth-cfg}
\end{table}


\section{COMPARISON TO DQN}
\label{appendix:dqn}

\begin{figure*}
	\vspace{.1in}
	\begin{center}
		\includegraphics[width=\linewidth, height=0.4\textheight]{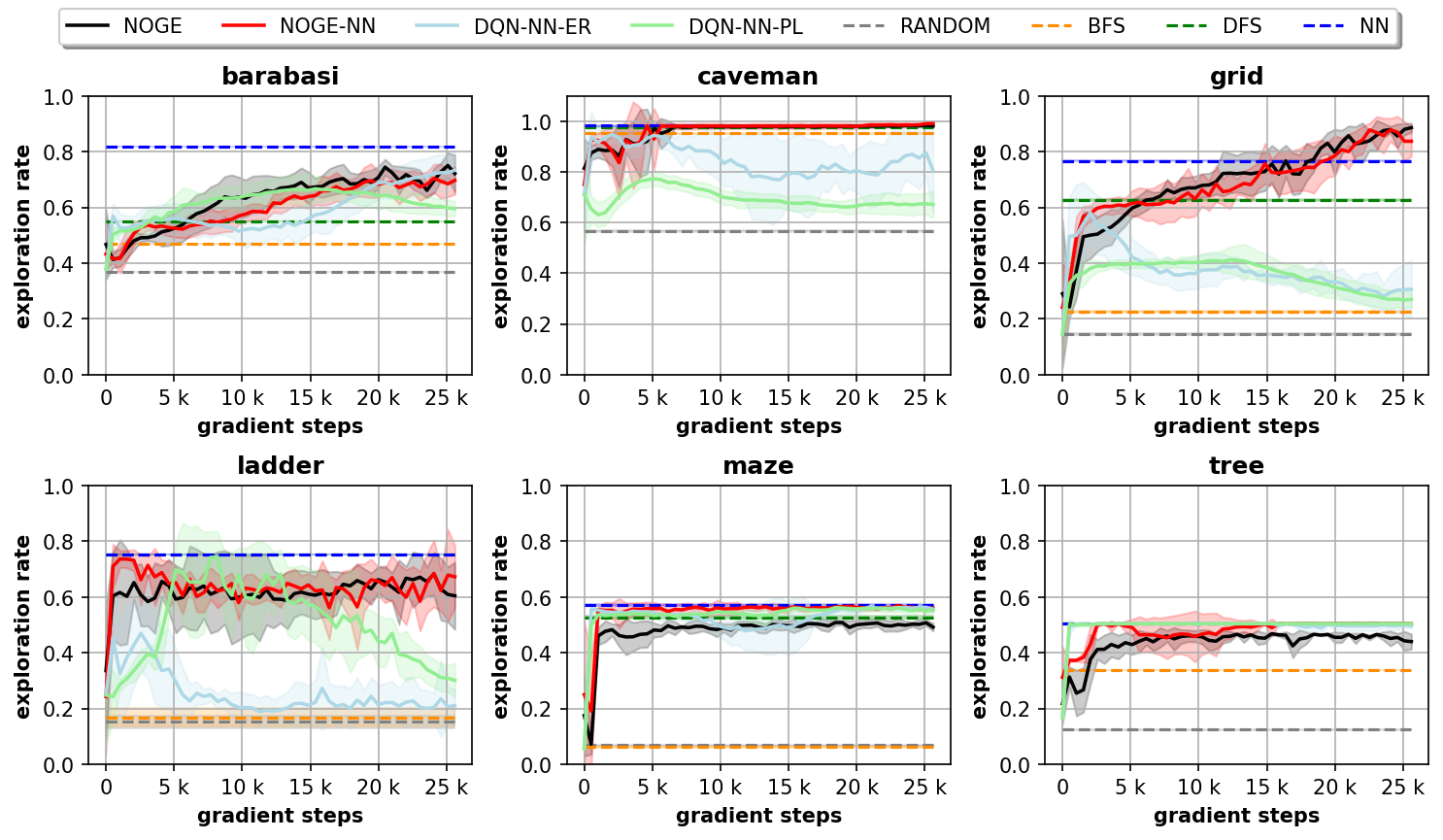}
		\caption{Exploration rate over gradient steps for the six procedurally generated data sets.}
		\label{fig:curves-synth-dqn}
	\end{center}
\end{figure*}

We originally tried using standard RL algorithms but found them to be unstable compared to DFP. In Figure~\ref{fig:curves-synth-dqn}, we additionally show the test exploration rate of a DQN \citep{mnih2015human} during training on the procedurally generated data sets. The curves show mean and standard deviation over 5 random seeds. Except for the original path length reward function (PL), described in the paper, we also tried using the exploration rate difference between two time steps (ER). In both cases the DQN had access to the NN feature. We can see that in 3 out of 6 data sets, the DQN struggles and degrades to solutions much worse than those found by NOGE.

%% file: tables/cfg_synth.tex
\begin{tabular}{lr}
\toprule
                            Parameter &                                 Value \\
\midrule
                       test set ratio &                                   0.2 \\
       max. episode steps ($T_{max})$ &                                   500 \\
                         node history &                                     2 \\
                        feature range &                           [-0.5, 0.5] \\
                 target normalization &                                True \\
                       training steps &                                 25600 \\
                  evaluation episodes &                                    50 \\
         env. steps per tr. step &                                    32 \\
          tr. steps per evaluation &                                   512 \\
 replay buffer size ($|\mathcal{R}|$) &                                 20000 \\
                  $\varepsilon_{max}$ &                                     1 \\
                  $\varepsilon_{min}$ &                                  0.15 \\
 temporal coefficients ($\mathbf{g}$) &  [0, 0, 0, $\frac{1}{4}$, $\frac{1}{4}$, $\frac{1}{2}$,$\frac{1}{2}$, 1] \\
                   minibatch size (B) &                                    32 \\
                        learning rate &                                0.0001 \\
\bottomrule
\end{tabular}